\documentclass[cis]{ipart_v1}

\usepackage{graphicx}
\usepackage{tikz}

\firstpage{213}
\Vol{14}
\Issue{4}
\Year{2014}

\usetikzlibrary{matrix,decorations.pathreplacing}

\theoremstyle{plain}

\theoremstyle{definition}

\theoremstyle{definition}

\begin{document}

\title[TransLoc3D]{TransLoc3D: Point Cloud based Large-Scale Place Recognition Using Adaptive Receptive Fields}

\author[T.-X.~Xu, Y.-C.~Guo, Z.~Li, G.~Yu, Y.-K.~Lai, and S.-H.~Zhang]{Tian-Xing Xu, Yuan-Chen Guo, Zhiqiang Li, Ge Yu, Yu-Kun Lai, and Song-Hai Zhang$^\ast$\blfootnote{* corresponding author}}

%

\begin{abstract}
Place recognition plays an essential role in the field of autonomous driving and robot navigation. Point cloud based methods mainly focus on extracting global descriptors from local features of point clouds. Despite having achieved promising results, existing solutions neglect the following aspects, which may cause performance degradation: (1) huge size difference between objects in outdoor scenes; (2) moving objects that are unrelated to place recognition; (3) long-range contextual information. We illustrate that the above aspects bring challenges to extracting discriminative global descriptors. To mitigate these problems, we propose a novel method named TransLoc3D, utilizing adaptive receptive fields with a point-wise reweighting scheme to handle objects of different sizes while suppressing noises, and an external transformer to capture long-range feature dependencies. As opposed to existing architectures which adopt fixed and limited receptive fields, our method benefits from size-adaptive receptive fields as well as global contextual information, and outperforms current state-of-the-arts with significant improvements on popular datasets.
\end{abstract}

\maketitle



\section{Introduction}
Navigation systems are essential for robots and self-driving cars to accurately localize themselves in complex outdoor scenes, which commonly depend on Global Positioning System (GPS). When GPS signal is not available, an alternative method is to sense, monitor, and gather the surrounding information of agents, such as the geometry of the buildings and roads, from depth sensors or RGB cameras, and then perform localization by recognizing the current place. 
Compared with point clouds obtained from depth sensors, images taken from RGB cameras are more sensitive to illumination changes, which may lead to significant performance degradation \cite{uy2018pointnetvlad}. To alleviate this problem, more and more works~\cite{uy2018pointnetvlad,zhang2019pcan,sun2020dagc,liu2019lpd,komorowski2021minkloc3d,zhou2021ndt} began to focus on place recognition based on 3D point clouds due to their inherent invariance to illumination.

\begin{figure}[!t]
\centering
\includegraphics[width=0.9\linewidth]{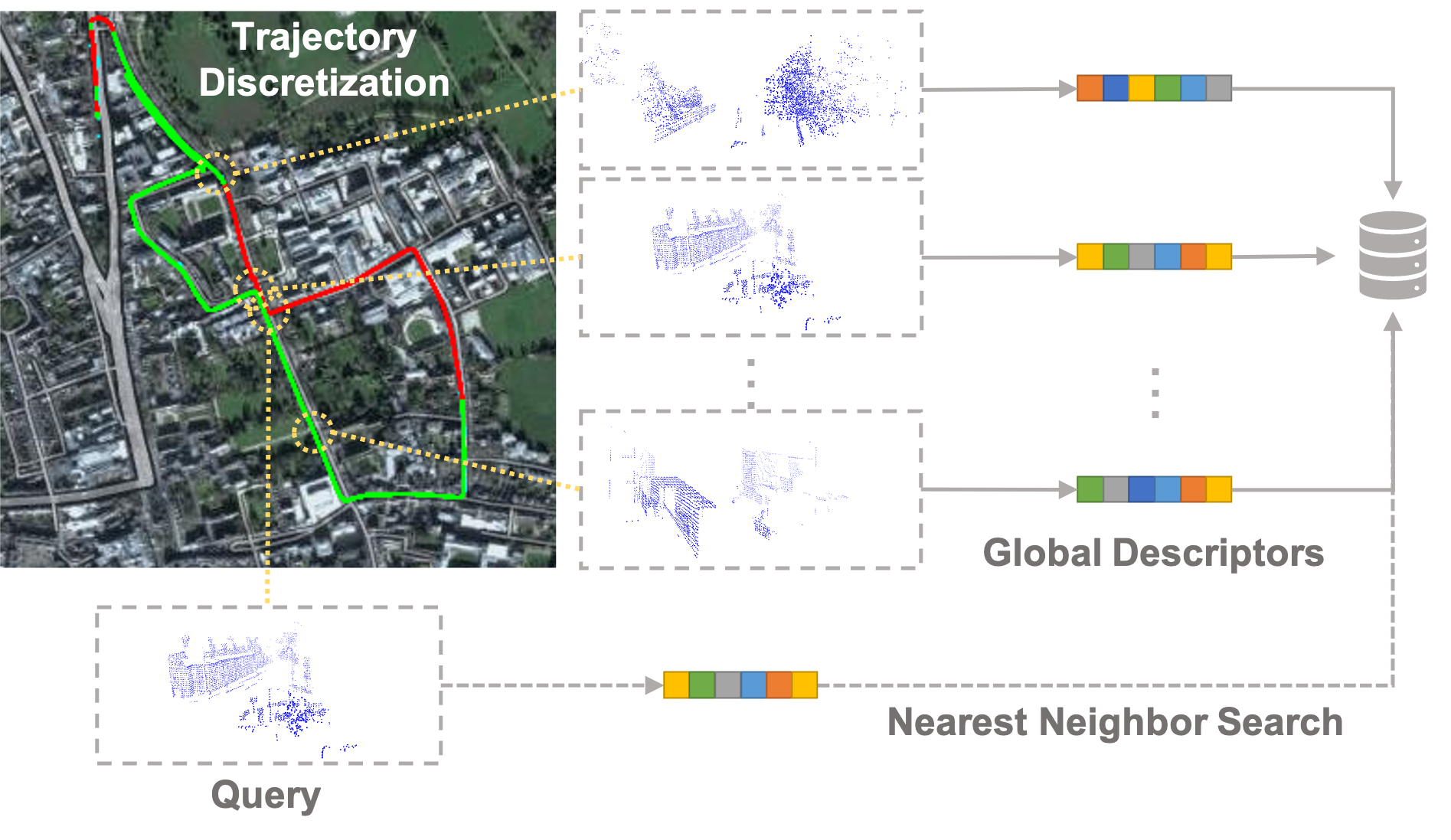}
\caption{Pipeline of point cloud based place recognition. The continuous trajectories are discretized into ``places" represented by the scanned point clouds. A recognition model first produces a discriminative descriptor for each point cloud, and then finds the closest match in existing point clouds for a query using the similarity of their descriptors.}
\label{fig:retrv}
\end{figure}

Large-scale point cloud based place recognition is often regarded as an instance retrieval problem, as illustrated in Fig.~\ref{fig:retrv}.
Although methods have been proposed to achieve promising results on research datasets, accurate and robust place recognition remains a challenging problem for the following reasons. First, in complex outdoor scenes, objects may differ drastically in size, whereas most existing methods perform feature extraction utilizing fixed receptive fields without consideration of size difference. For small objects like vehicles, large receptive fields will capture unrelated information, making the extracted features less discriminative, while for large objects such as buildings, small receptive fields would fail to encode the complete geometric structure. Second, moving objects in the scene, like pedestrians, are not related to place recognition, which requires the feature extraction process to be robust to such noise.
Third, most of the existing methods only consider extracting features of local regions, while neglecting long-range contextual information. We argue that the lack of long-range contextual information limits the representation power of the descriptors. 
To address the above issues, we propose a new architecture named TransLoc3D. Our architecture first processes the input points using sparse voxelization and 3D sparse convolution, followed by a novel feature extraction pipeline, and produces global descriptors by NetVLAD~\cite{arandjelovic2016netvlad,uy2018pointnetvlad}. 
Our proposed feature extraction pipeline is capable of adaptively adjusting receptive field sizes in accordance with the targeted objects, which utilizes a point-wise feature reweighting scheme to reweight features of multiple receptive scales by a learned attention map. 
After that, we adopt external attention layers~\cite{guo2021beyond} to capture long-range contextual information. Combining the strengths of 
adaptive receptive fields and a transformer-based architecture,
TransLoc3D can produce more discriminative global descriptors for point clouds. Quantitative results show that TransLoc3D surpasses existing methods and achieves state-of-the-art average recall on widely adopted benchmarks. We also demonstrate the ability of TransLoc3D to alleviate the above issues by qualitative visualizations.

Our contributions can be summarized as:
\begin{itemize}
\item We 
argue that 
taking object size differences into consideration is necessary for point cloud based recognition of complex scenes, and propose to use adaptive receptive fields for feature extraction. A point-wise reweighting scheme is used to fuse features from different scales and suppress noises.
\item We design a new architecture named TransLoc3D, which effectively combines the advantages of 
adaptive receptive fields and transformer, making it suitable for the place recognition task. We also provide a qualitative analysis of these modules by visualizing the results.

\item Extensive experiments demonstrate that the proposed TransLoc3D achieves state-of-the-art results on four popular benchmarks, namely Oxford RobotCar, B.D., U.S. and R.A. datasets. 

\end{itemize}


\section{Related Work}

\subsection{3D Point Cloud Based Place Recognition}


PointNetVLAD~\cite{uy2018pointnetvlad} is the first learning-based method for large-scale place recognition. It follows the design of PointNet~\cite{qi2017pointnet} to extract point-wise features and then adopts NetVLAD~\cite{arandjelovic2016netvlad}
to transform point-wise features into a global discriminative descriptor. 
Following PointNetVLAD, PCAN~\cite{zhang2019pcan} incorporates a Point Contextual Attention module into the PointNet architecture, which can predict the significance of each independent point feature based on contextual information. However, both of these methods ignore the spatial point distribution in local areas, which limits the representation power of the global descriptors.
To capture local geometry information, LPD-Net~\cite{liu2019lpd} adopts a graph-based aggregation module in both feature space and Cartesian space and achieves state-of-the-art performance. 
Previously mentioned works all operate directly on unordered point sets. In contrast, MinkLoc3D~\cite{komorowski2021minkloc3d} and Minkloc++~\cite{komorowski2021minkloc++} use an alternative data representation for place recognition. Point clouds are first voxelized into a sparse voxel representation, and then a 3D sparse Convolution Neural Network (CNN) built on a Feature Pyramid Network~\cite{lin2017feature} is adopted to extract informative local features. 
However, simply stacking convolution layers may ignore long-range contextual information, and conventional CNNs with fixed receptive fields fail to tackle the size difference problem.

\subsection{Transformers in Computer Vision}

Recently, inspired by the success of Vision Transformer~\cite{dosovitskiy2021an}, more and more researchers focus their attention on applying the Transformer architecture to vision tasks~\cite{han2021tit,wang2021pyramid,dai2020up,wang2020end,ham,zheng2020rethinking}.
NDT-Transformer~\cite{zhou2021ndt} is the first deep learning architecture modeled upon a standard Transformer for place recognition. In this model, each point cloud is first transformed into the Normal Distribution Transform Cell (NDT Cell) representation~\cite{magnusson2007scan}, and then fed to a transformer with 3 encoders to capture long-range contextual information. However, transformer-based models suffer from large memory consumption, which limits the batch size for deep metric learning and further influences the performance of models. 

\subsection{Multi-Scale Receptive Fields}
Numerous experiments~\cite{nelson1978orientation,pettet1992dynamic,sceniak1999contrast} in neuroscience have suggested that the receptive field size of a neuron is not fixed but adaptive to the input bioelectric signals. However, this property does not receive sufficient attention in constructing CNNs. 
InceptionNetV1~\cite{SzegedyLJSRAEVR2015googlenet} is the first architecture aggregating multi-scale information within the same layer via a simple concatenation mechanism. 
Following InceptionNet, many methods try to improve feature representation by concatenating multi-scale features, but they often fail to select appropriate scales for the targeted objects.
The follow-up work SKNet~\cite{li2019selective} enhances this architecture using an attention mechanism to fuse multi-scale information from different receptive fields. Compared with simple concatenation, 
the attention mechanism is more suited to adaptively adjusting the receptive field sizes based on input
and has potential for tackling the size difference problem in outdoor scenes. 

\section{Method}

\begin{figure}
\begin{center}
   \includegraphics[width=1.0\linewidth]{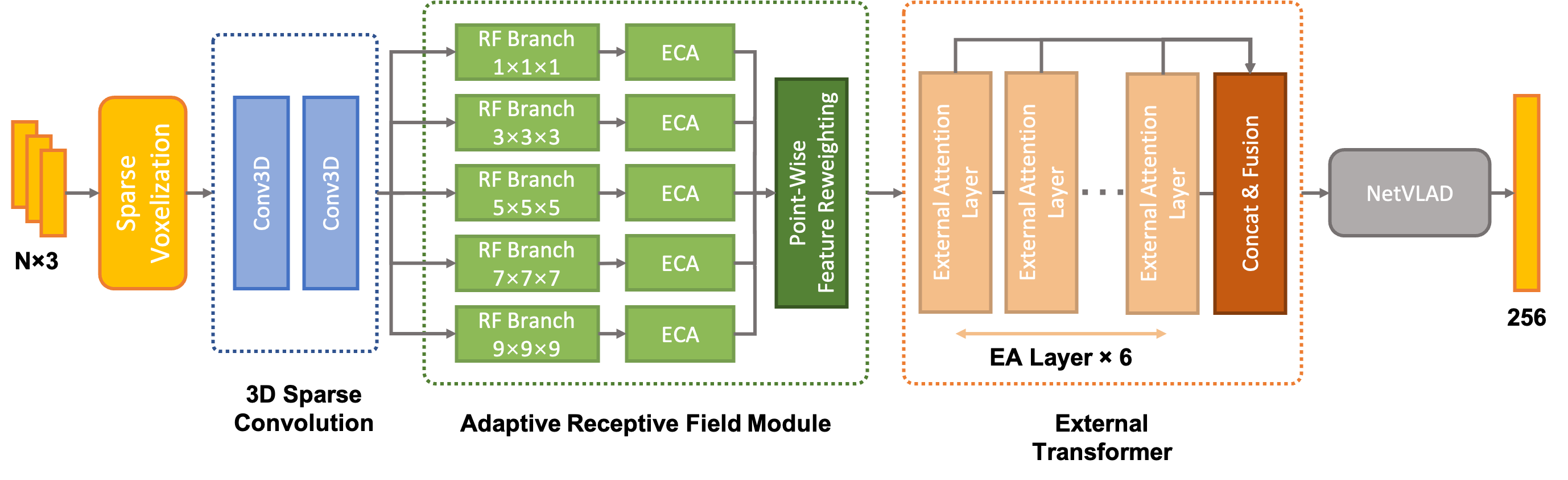}
\end{center}
   \caption{TransLoc3D Network Architecture. Our proposed TransLoc3D consists of four parts stacked in series, including a 3D sparse convolution module, an adaptive receptive field module with point-wise feature reweighting, an external transformer and a NetVLAD module. Each ``RF Branch" in the adaptive receptive field module consists of different numbers of convolutional blocks to extract features of different receptive field sizes, while the module is capable of adaptively adjusting the size of its receptive field according to the input point cloud. }
\label{fig:transloc3d}
\end{figure}

\subsection{Overview}

As illustrated in Fig.~\ref{fig:transloc3d}, our proposed TransLoc3D includes four main parts, a 3D sparse convolution module, an adaptive receptive field module with point-wise feature reweighting, an external transformer and a NetVLAD \cite{arandjelovic2016netvlad} module. We use triplet margin loss~\cite{HermansBL17tripletloss} with batch hard mining~\cite{wu2017sampling} to train our network, which requires a larger batch size to find more informative triplets. Instead of using raw point clouds for feature extraction~\cite{qi2017pointnet,qi2017pointnet++,liu2019relation}, which requires quadratic space complexity to compute the neighborhood of each point, sparse voxel representation enables our network to obtain the neighborhood of each voxel using a hash algorithm with a linear space complexity. Therefore, we transform the input point cloud into a sparse voxel representation and adopt 3D Sparse Convolution (Sp-Conv)~\cite{choy20194d} as a basic unit to build our network. 

We first employ a small network with two Sp-Conv layers to aggregate local geometric information and the details can be found in the appendix.
Then features of different receptive field sizes are extracted, and a point-wise reweighting scheme is designed to adaptively fuse these features, while being able to suppress noises. 
Next, we introduce a transformer architecture to capture long-range contextual information.
Finally, we adopt a NetVLAD \cite{arandjelovic2016netvlad} layer to aggregate local features of each voxel to produce a global descriptor for recognition. NetVLAD learns $K$ cluster centers and sums the difference between the local descriptors and the corresponding cluster centers, to obtain a permutation-invariant descriptor. To make fair comparisons, NetVLAD is followed by a Multi-Layer Perceptron (MLP) to produce a descriptor of the same size as that of previous works.

\subsection{Adaptive Receptive Field Module with Point-Wise Feature Reweighting}

Although a number of methods have obtained outstanding performance, place recognition remains challenging due to the existence of unrelated moving objects and the effects of objects in different sizes. We argue that fixed receptive fields cannot well tackle huge size differences between objects. To capture clean and consistent geometric information of objects in all sizes, we propose a novel module capable of adaptively adjusting the receptive field size with a point-wise reweighting scheme.


\begin{figure}
\begin{center}
   \includegraphics[width=0.9\linewidth]{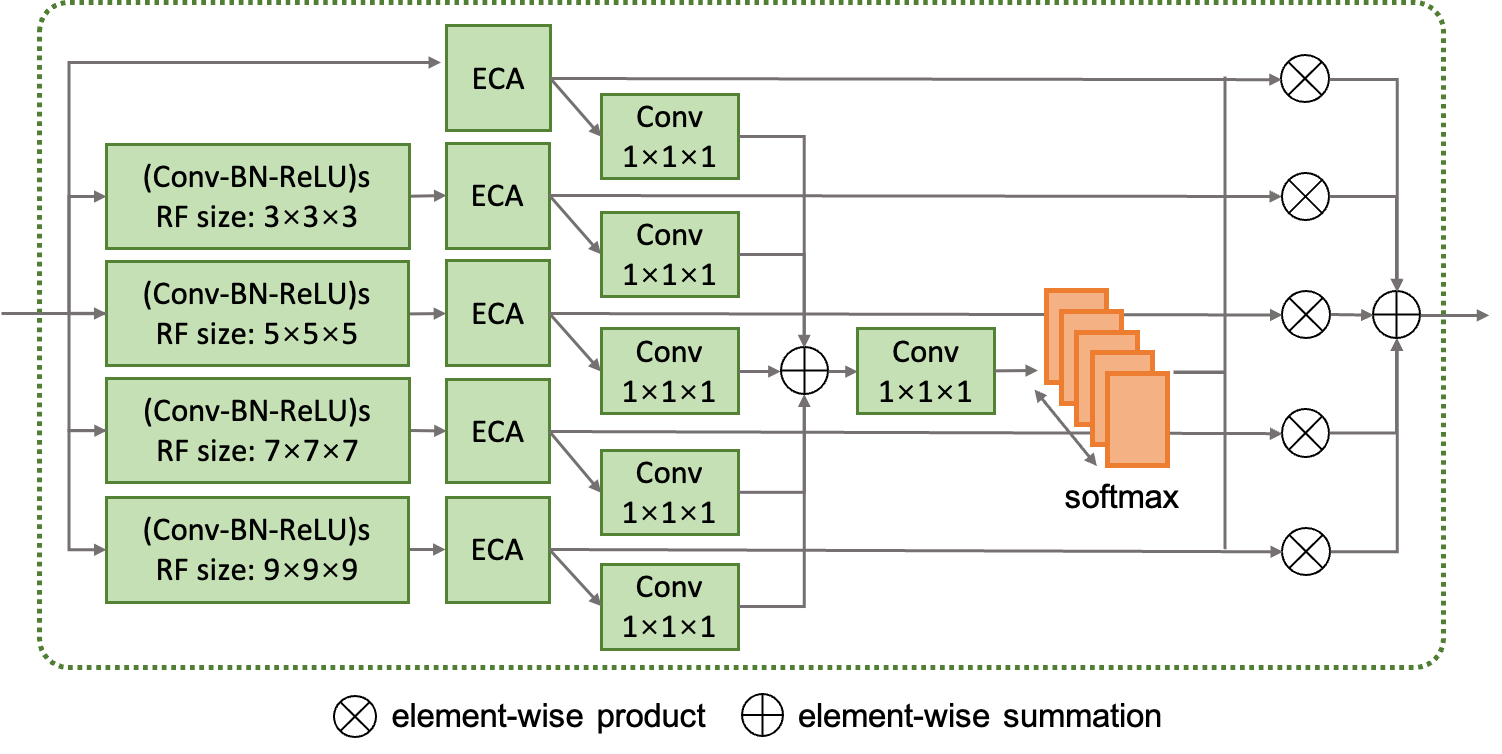}
\end{center}
   \caption{Adaptive receptive field module with a point-wise reweighting scheme. Information from neighborhoods of different sizes is aggregated on each branch, and then fed to ECA modules~\cite{wang2020eca} for further enhancement. A point-wise reweighting scheme is adopted to fuse information from different branches.}
\label{fig:arfm}
\end{figure}

The design of our proposed module is inspired by Selective Kernel Convolution (SK-Conv) \cite{li2019selective}. Unlike SK-Conv, we replace the dilated convolutions \cite{chen2017deeplab} with conventional convolutions because the combination of dilated convolution and sparse feature maps leads to significant performance degradation in our experiments. We also replace the lightweight fusion mechanism with a novel point-wise reweighting scheme, based on our observation that objects of the same category still have different sizes. Formally, for the given feature map $X \in \mathbb{R}^{H\times W\times D\times C}$ output by the 3D sparse convolution module, we first conduct transformations $\mathcal{F}_i: X\rightarrow X'_i\in \mathbb{R}^{H\times W\times D\times C}$ with different receptive field sizes. Fig.~\ref{fig:arfm} illustrates that our module consists of five parallel branches. For computing efficiency, branches with receptive field sizes larger than $5\times5\times5$ is implemented by stacking convolution layers with kernel size $3\times3\times3$ and $5\times5\times5$. All convolutions are followed by a batch normalization \cite{ioffe2015batch} layer and ReLU activation, except for the last layer of each branch.

Considering that large receptive field would inevitably capture small objects, we employ an ECA (Efficient Channel Attention) \cite{wang2020eca} module to suppress noisy features. Formally, for the $i$-th branch, we multiply the feature map $X'_i\in \mathbb{R}^{H\times W\times D\times C}$ by a channel-wise weighting vector $w'\in \mathbb{R}^C$
\begin{gather}
w_i' = \sigma(\phi_i(\text{AvgPool}(X'_i))) \\
X''_i = X'_i \cdot w_i'
\end{gather}
where $\sigma$ denotes the Sigmoid function and $\phi_i$ indicates 1D convolution with a kernel of size $k$ 
along the channel dimension to model local cross-channel interactions. The hyperparameter $k$ can be adaptively determined by channel dimension $C$ as proposed in ECA\cite{wang2020eca}. In the last step, the information originated from multiple branches is fused together by a point-wise reweighting scheme
\begin{equation}
X_\text{out} = \sum_{i} W_i'' \cdot X_i''
\end{equation}
where $W_i'' \in \mathbb{R}^{H\times W\times D\times C}$ denotes the attention weighting map of the $i$-th branch.
As illustrated in Fig.~\ref{fig:arfm}, to obtain the weighting maps $W_i''$, we fuse results from multiple branches via an element-wise summation first. 
\begin{equation}
T = \sum_{i}\delta_i(X_i'')
\end{equation}
Here $\delta_i$ is defined as a non-linear mapping function on the $i$-th branch, implemented by stacking $1\times1$ convolutions with batch normalization \cite{ioffe2015batch} layers and ReLU non-linear functions. 
Notably, Selective Kernel Convolution \cite{li2019selective} squeezes the global spatial information for higher efficiency while we preserve the spatial dimensions. Taking buildings as an example, we observe that regions representing buildings have huge size differences due to occlusions, thus our intuition is that the weighting vectors should be different for different positions. 
Therefore the $i$-th weighting map $W_i''$ 
is defined as
\begin{equation}
W_i'' = \frac{\exp(\varphi_i(T))}{\sum_j \exp(\varphi_j(T))}
\end{equation}
where $\varphi_i$ denotes an element-wise non-linear mapping from aggregated information to the weighting maps of the $i$-th branch. In practice, it is implemented by a convolution with $C$ kernels of shape $1\times1\times 1$, followed by batch normalization.

\subsection{External Transformer}

As mentioned before, the neglect of long-range contextual information may limit the representation power of global descriptors. Therefore, we adopt an External Transformer~\cite{guo2021beyond} to aggregate information from both nearby and far-away voxels in spatial dimensions due to its linear space complexity and high computational efficiency. As illustrated in Fig.~\ref{fig:transloc3d}, the transformer includes 6 External Attention (EA) layers stacked in series, which can be written as follows
\begin{gather}
\text{EA}(Q_{i},K_{i}^{(M)},V_{i}^{(M)}) = \text{Softmax}(\frac{Q_{i}{K_{i}^{(M)}}^{T}}{\sqrt{d_k}})V_{i}^{(M)}
\end{gather}
where $K_{i}^{(M)}, V_{i}^{(M)}\in \mathbb{R}^{S\times d}$ denote the $i$-th head of two external learnable memory units and the hyper-parameter $S$ is the number of keys and values in the EA mechanism. For further reduction of model parameters, the values $V_i^{(M)}$ are obtained by applying a linear mapping $\phi$ to the keys $K_i^{(M)}$ instead of an extra memory unit
\begin{equation}
V_i^{(M)} = \phi(K_i^{(M)})
\end{equation}

The space complexity of External Attention~\cite{guo2021beyond} is $O(N\times S)$, thus we can control the amount of memory consumed in  the training process by adjusting the hyper-parameter $S$. Same as PCT~\cite{guo2020pct}, we incorporate an offset-attention module with a small modification\footnote{We replace the input to the LBR network $F_\text{in} - F_\text{EA}$ proposed in PCT~\cite{guo2020pct} with $F_\text{EA} - F_\text{in}$, which does not affect the representation ability of the model in theory, but has a slight improvement in our experiments. We assume this is because the LBR network now needs to learn a mapping closer to identity, which is easier to model. } 
to External Attention layer for further enhancement
\begin{align}
F_\text{out} &= \text{LBR}(F_\text{EA} - F_\text{in})+F_\text{in} \\
F_\text{EA} &= [\text{EA}(Q_{i},K_{i}^{(M)},V_{i}^{(M)})] \notag\\ 
&= [\text{Softmax}(Q_{i}{K_{i}^{(M)}}^{T})V_{i}^{(M)}]
\end{align}
Here $\text{LBR}$ combines {\em Linear}, {\em BatchNorm} and {\em ReLU} layers, $F_\text{in}, F_\text{out}$ are the input and output features of the EA module, and $[ \cdot ]$ denotes the concatenation operation. Finally, the output of each EA layer is concatenated along the channel dimension, followed by a transformation to aggregate information from different levels. 


\subsection{Network Training}
Although a number of sophisticated loss functions have been proposed in deep metric learning, recent works \cite{musgrave2020metric,roth2020revisiting} show that their advantages over the classical triplet margin loss \cite{HermansBL17tripletloss} are moderate. We use triplet margin loss to train our network, which requires an anchor, a positive example (structurally similar to the anchor) and a negative example (structurally dissimilar to the anchor):
\begin{equation}
\mathcal{L}_\text{triplet} = \frac{1}{N}\sum_{i=1}^N [||\delta_\text{a}-\delta_\text{p}||_2-||\delta_\text{a}-\delta_\text{n}||_2+\alpha]_+    
\end{equation}
Here $N$ is the number of training samples in a batch, $\delta_\text{a}$, $\delta_\text{p}$ and $\delta_\text{n}$ denote the global descriptors of the anchor, positive and negative point clouds respectively (the index $i$ that refers to a specific sample is omitted to avoid clutter).
$[\cdot]_+$ denotes the function $\max(\cdot, 0)$ and $\alpha$ is the constant margin. 
Same as Minkloc3D \cite{komorowski2021minkloc3d}, at the beginning of each epoch the training set is partitioned into batches by randomly sampling positive pairs from the remaining data repeatedly. For each batch we compute two $N\times N$
binary masks indicating the structural similarity between each pair of point clouds. 
The discriminative descriptor of each point cloud in a batch is obtained by our proposed network. Then we construct informative triplets via the batch hard mining approach \cite{wu2017sampling} using the two binary masks.    

At the early stage of training, the model cannot produce sufficiently discriminative descriptors and mode collapse is more likely to occur with a large batch size. Therefore, we adaptively adjust the batch size as the training continues, as is proposed in Minkloc3D \cite{komorowski2021minkloc3d}. If the average number of triplets producing non-zero loss accounts for over $\eta$ of the total number, the batch size will be enlarged to $\tau$ 
times of the previous value as long as the batch size is still below a maximum threshold. Here $\tau$ is a hyper-parameter larger than 1.

\section{Experiments}

\subsection{Datasets}


We use a modified Oxford RobotCar dataset \cite{RobotCarDatasetIJRR} and three 
other datasets introduced in PointNetVLAD~\cite{uy2018pointnetvlad}, including Business District (B.D.), Residual Area (R.A.) and University Sector (U.S.) to evaluate our method. Point clouds in Oxford RobotCar dataset are obtained by a Sick LMS-151 2D LiDAR scanner mounted on a moving vehicle, while others are obtained from a Velodyne-64 LiDAR scanner. 
The places are sampled with a fixed interval on the continuous trajectory of the vehicle, and the corresponding point clouds are generated by dividing the global map into a set of submaps. During training, point cloud pairs with a distance less than 10m are defined as positive pairs, while more than 50m are defined as negative pairs. The rest of point cloud pairs are neither positive nor negative. To better learn geometric features, the non-informative points on the ground are removed, then the number of points is uniformly downsampled to 4096. Coordinates of each point are shifted and scaled to $[-1,1]$.

\subsection{Implementation Details}


To reduce the risk of overfitting, data augmentation is introduced into the preprocessing stage. Specifically, we adopt random jittering with a noise sampled from the normal distribution $\mathcal{N}(0, 0.001)$ and clipped to $[-0.002,0.002]$, random translation with an offset vector sampled from the uniform distribution $[-0.01, 0.01]^3$, random point removal with a probability of $d_r\sim [0.0,0.1]$, random symmetrical transformation and random rotation. We also use random fronto-parallel cuboid erasing approach proposed in \cite{komorowski2021minkloc3d} for further augmentation.  


Augmented point clouds are quantized with quantization step 0.01, and then fed to the network implemented by Minkowski Engine \cite{choy20194d}.
The hyperparameter $S$ of each EA layer is set to 256, 128, 128, 64, 64, 64 in sequence to capture rich local geometric information. The number of heads is set to 2. The concatenation of outputs from different attention layers is transformed to 512-dimensional space, and then fed to NetVLAD \cite{arandjelovic2016netvlad}. With regard to the hyperparameters within NetVLAD, the size of the cluster is set to 64 and the dimension of the output descriptor is set to 256 for fair comparisons. We also introduce context gating mechanism \cite{Antoine2017ContextGating}, which is initially proposed for large-scale video understanding, into NetVLAD to produce more informative descriptors.


We adopt Adam optimizer with an initial learning rate $2\times 10^{-4}$, and multiplied by $0.1$ on epoch 80, 120 and 160. The triplet loss margin $\alpha$ is set to $0.2$ in our experiments. The batch size is initially set to 32 and increased by 40\% once the number of active triplets is less than 70\% of the total. All experiments are conducted on a server with 6 NVidia GeForce GTX 1080Ti GPUs and an Intel i7-6850K CPU.

\subsection{Quantitative Comparisons}

\begin{table}
\begin{center}
\begin{tabular}{lcccc}
\hline
Method & Oxford & B.D. & R.A. & U.S.\\
\hline
PointNetVLAD & 80.3 & 65.3 & 60.3 & 72.6 \\
PCAN & 83.8 & 66.8 & 71.2 & 79.1 \\
DAGC & 87.5 & 71.2 & 75.7 & 83.5 \\
LPD-Net & 94.9 & {\bf 89.1} & 90.5 & {\bf 96.0} \\
SOE-Net & 96.4 & 88.5 & {\bf 91.5} & 93.2 \\
Minkloc3D & 97.9 & 88.5 & 91.2 & 95.0 \\
NDT-Transformer & 97.7 & - & - & - \\
Minkloc++ & 98.2 & 82.7 & 85.1 & 93.0 \\
TransLoc3D(ours) & {\bf 98.5} & 88.4 & {\bf 91.5} & 94.9 \\

\hline
\end{tabular}
\end{center}
\caption{Evaluation Results({\em AR@1\%}). All the methods are trained on Oxford RobotCar dataset and evaluated on four datasets without finetuning. Our method achieves the state-of-the-art performance on Oxford RobotCar dataset, and has a competitive generalization ability. }
\label{tab:baseline1}
\end{table}
Following the same evaluation protocol introduced in PointNetVLAD~\cite{uy2018pointnetvlad},
we compare our model with previous works and the results are shown in Tab.~\ref{tab:baseline1}. All these models are trained on modified Oxford RobotCar dataset and evaluated on test splits of four datasets without further finetuning. It is worth noting that Minkloc++ is a multimodal architecture taking geometric information from RGB images and 3D point clouds as input and we only use the 3D modality sub-network. If the results are not reported in the original paper, we run the evaluation by ourselves with publicly available source codes, otherwise we take the results reported by authors with the identical evaluation protocol.

Our proposed TransLoc3D achieves state-of-the-art results on Oxford RobotCar dataset, with a 0.3\% higher average recall@1\% than the runner-up method, Minkloc++.
Compared with NDT-Transformer which is also based on a Transformer architecture, our method adopts a nearly cost-free preprocessing step and achieves a remarkable improvement of 0.8\% on average recall@1\%, which shows the significance of adaptive receptive fields. For generalization capability, our model surpasses other models on the R.A. dataset while is slightly worse than LPD-Net on other datasets. We augue that there exist huge differences in data distribution and LPD-Net enhances the input with hand-crafted features, which improves the generalization ability of the network by introducing prior knowledge. The following experiment shows that our method surpasses LPD-Net after finetuning.


\begin{table}
\begin{center}
\begin{tabular}{lcc}
\hline
Method & {\em AR@1} & {\em AR@1\%}\\
\hline
PointNetVLAD & 63.3 & 80.3  \\
PCAN & 70.7 & 83.8\\
DAGC & 73.3 & 87.5 \\
LPD-Net & 86.3 & 94.9\\
SOE-Net & 89.4 & 96.4\\
Minkloc3D & 93.8 & 97.9 \\
NDT-Transformer & 93.8 & 97.7 \\
Minkloc++ & 93.9 & 98.2\\
TransLoc3D(ours) & {\bf 95.0} & {\bf 98.5} \\

\hline
\end{tabular}
\end{center}
\caption{Evaluation Results on Oxford RobotCar dataset. Our method has a remarkable improvement of 1.1\% on {\em AR@1} and 0.3\% on \em{AR@1\%}. }
\label{tab:baseline2}
\end{table}

\begin{table}
\begin{center}
\resizebox{\linewidth}{!}{
\begin{tabular}{lcccccc}
\hline
Method & \multicolumn{2}{c}{B.D.} & \multicolumn{2}{c}{R.A.} & \multicolumn{2}{c}{U.S.} \\
 &{\em AR@1} & {\em AR@1\%}&{\em AR@1} & {\em AR@1\%}&{\em AR@1} & {\em AR@1\%}\\
\hline
PointNetVLAD~\cite{uy2018pointnetvlad} & 80.1 & 86.5 & 82.7 & 93.1 & 86.1 & 90.1  \\
PCAN~\cite{zhang2019pcan} & 80.3 & 87.0 & 82.3 & 92.3 & 83.7 & 94.1 \\
DAGC~\cite{sun2020dagc} & 81.3 & 88.5 & 82.8 & 93.4 & 86.3 & 94.3 \\
LPD-Net~\cite{liu2019lpd} & 90.8 & 94.4 & 90.8 & 96.4 & 94.4 & 98.9 \\
SOE-Net~\cite{xia2020soe} & 89.0 & 92.6 & 90.2 & 95.9 & 91.8 & 97.7 \\
Minkloc3D~\cite{komorowski2021minkloc3d} & 94.0 & 96.7 & 96.7 & 99.3 & 97.2 & 99.7 \\
NDT-Transformer~\cite{zhou2021ndt} & - & - & - & - & - & -\\
Minkloc++~\cite{komorowski2021minkloc++} & 91.8 & 95.5 & 95.3 & 98.5 & 96.5 & 99.5 \\
TransLoc3D(ours) & {\bf 94.8} & {\bf 97.4} & {\bf 97.3} & {\bf 99.7} & {\bf 97.5} & {\bf 99.8} \\

\hline
\end{tabular}}
\end{center}
\caption{Evaluation Results. Place recognition methods are trained on Oxford RobotCar, U.S. and R.A. datasets. After finetuning, our method surpasses other methods with a significant improvement of average recall on all datasets.}
\label{tab:baseline3}
\end{table}

Due to that {\em AR@1\%} is already close to 100\%, we also compare the performance of various models using {\em AR@1} on the Oxford RobotCar dataset. As illustrated in Tab.~\ref{tab:baseline2}, our method is significantly better than the previous state-of-the-art model, with an improvement of 1.1\%, which shows that TransLoc3D has stronger discrimination ability than other methods. 

We also test the performance with finetuning. Same as PointNetVLAD~\cite{uy2018pointnetvlad}, the training subsets of U.S. and R.A. are also added to the training data 
in addition to the Oxford RobotCar dataset, in order to verify the generalization ability of models on unseen scenarios. Inspired by transfer learning, we pretrain our model on Oxford RobotCar dataset, and then finetune the model with a small learning rate on downstream datasets. Tab.~\ref{tab:baseline3} illustrates the evaluation results of different methods under this setting, where the evaluation of Minkloc++~\cite{komorowski2021minkloc++} is conducted by ourselves due to the lack of reported results, and other results are reported by the authors. It is shown that our method is superior to previous methods with a remarkable improvement on three datasets. 
The comparison to LPD-Net~\cite{liu2019lpd} shows that hand-crafted features contribute greatly to the model performance under the condition that the amount of data is relatively small or there exist huge differences in data distribution. Otherwise, learning-based models have a stronger representation power.  

\subsection{Visualization}

\begin{figure}[t]
\begin{center}
   \includegraphics[width=0.8\linewidth]{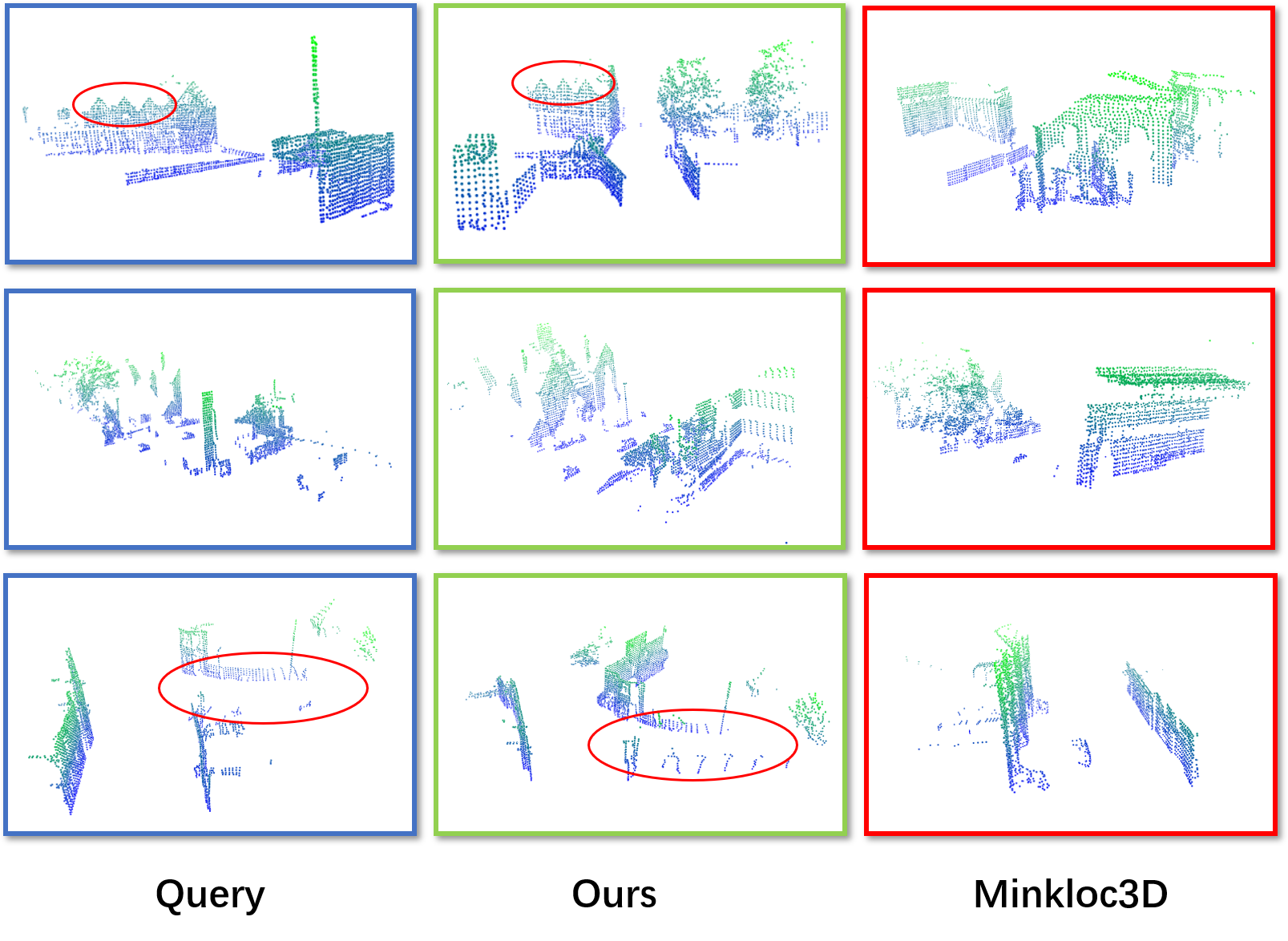}
\end{center}
   \caption{Visualization of some representative retrieval results. Our method recognize the correct places while Minkloc3D~\cite{komorowski2021minkloc3d} fails under these cases. }
\label{fig:visual}
\end{figure}

To intuitively explain how our model addresses three issues mentioned before, we visualize the retrieval results and perform qualitative comparisons,  
which is illustrated in Fig.~\ref{fig:visual}. The leftmost column shows the query point cloud and other columns show its nearest neighbors using our method and Minkloc3D. In the first case, the large receptive field of Minkloc3D neglects peaked roofs (red circle) of the building, while our model can adaptively adjust the receptive field size according to the target object size. In the second case, points distributed on the continuous building surfaces are divided into several patches due to the occlusions brought in by pedestrians and trees, while our model is more robust than Minkloc3D under this circumstance. In the third case, the model is required to capture geometric information of the horizontal road (red circle) to produce a discriminative descriptor. Minkloc3D fails to encode the road due to its limited receptive field size, while our transformer-based architecture enables our model to capture global information.    

\subsection{Ablation Studies}

\label{section:ablation}

\begin{table}
\begin{center}
\begin{tabular}{lcc}
\hline
Network & {\em AR@1} & {\em AR@1\%}\\
\hline
w/o adaptive receptive fields & 83.6 & 93.9 \\
w/o ECA & 94.5 & 98.4 \\
w/o point-wise feature reweighting & 94.6 & {\bf 98.5} \\
w/o transformer & 94.2 & 98.2 \\
ours full & {\bf 95.0} & {\bf 98.5} \\
\hline
\end{tabular}
\end{center}
\caption{Ablation study on several design choices. The elimination of each module leads to a significant degradation on the performance. }
\label{tab:ablation1}
\end{table}


We conduct ablation studies to evaluate the impact of different design choices of our method. In all experiments of this section, the network is trained and evaluated only using the Oxford RobotCar dataset. We first eliminate several independent modules including adaptive receptive field module, ECA~\cite{wang2020eca}, point-wise feature reweighting and transformer, referred as ``w/o adaptive receptive fields", ``w/o ECA", ``w/o point-wise feature reweighting" and ``w/o transformer". We denote the complete network architecture as ``ours full". The results are illustrated in Tab.~\ref{tab:ablation1}. The elimination of point-wise feature reweighting scheme is implemented by replacing weighted summation based on attention weights with a concatenation operation along the channel dimension, followed by a $1\times1\times1$ convolution used to aggregate information from different branches. Compared with the complete TransLoc3D model, other alternatives all have significant degradation on average recall@1, from 0.4\% to 11.4\%.

\begin{figure}[t]
\begin{center}
   \includegraphics[width=0.8\linewidth]{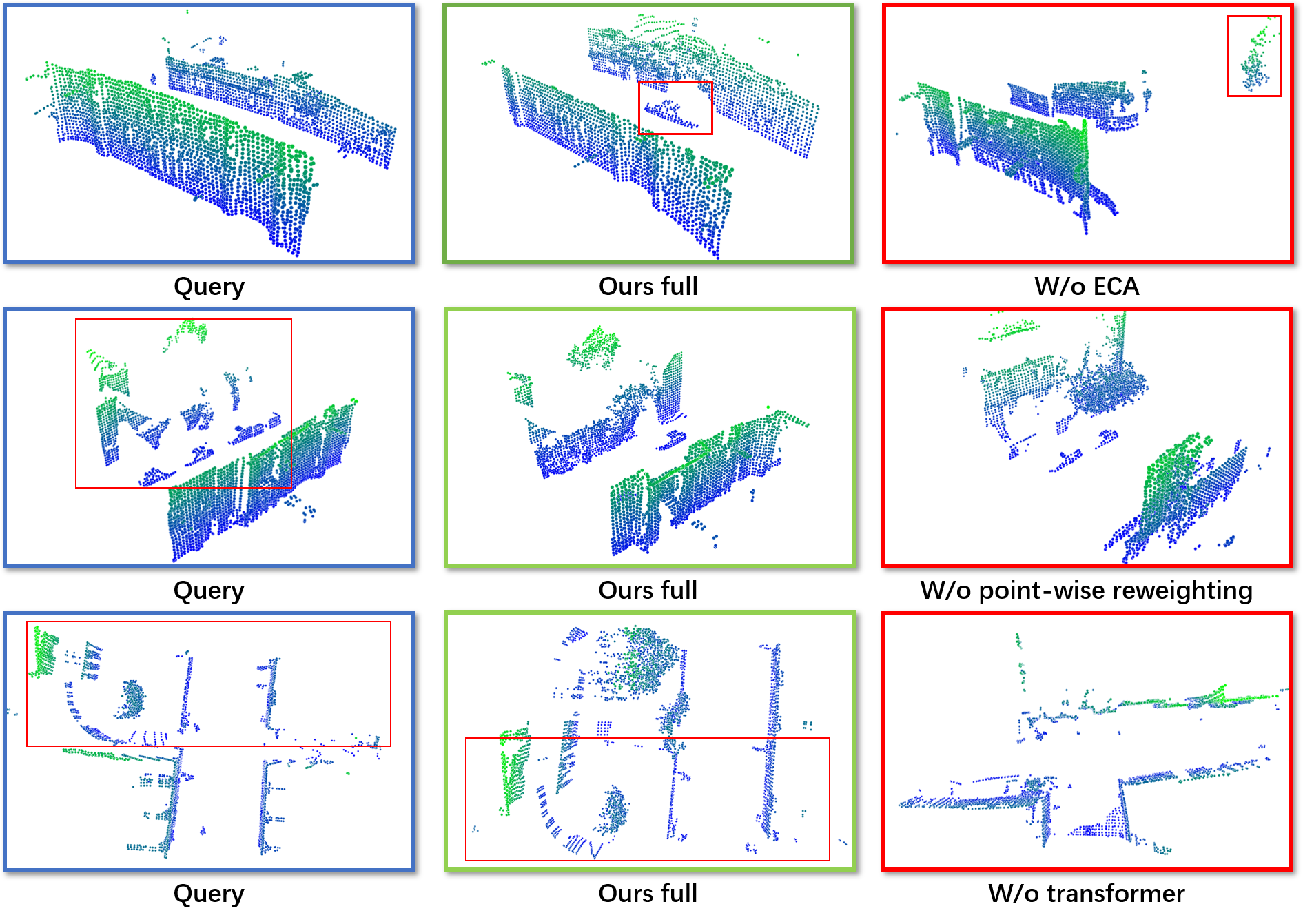}
\end{center}
   \caption{Visualization of some representative retrieval results. Our complete model recognizes correct places while others fail under these cases. }
\label{fig:visualx}
\end{figure}

To further explore the significance of these module, we visualize several typical cases in which the complete TransLoc3D model surpasses other models. As illustrated in Fig.~\ref{fig:visualx}, in the first case, the retrieval results are perturbed by noise including vehicles on the road and buildings in the distance. The ECA module enables our complete model to be more robust to noises than ``w/o ECA". In the second cases, the vehicles on the street obscures the huge building, which further changes the distribution of points and splits the whole point cloud into patches of different sizes. The point-wise reweighting mechanisms can adaptively adjusting the receptive field sizes according to the input point clouds, which is suitable for this case. In the third case, only part of the buildings (bottom of the retrieval result) can match the query point cloud, thus it requires long-range contextual information to produce sufficiently discriminative descriptors when local regions have similar geometry.

\begin{figure}[t]
\begin{center}
  \includegraphics[width=0.3\linewidth]{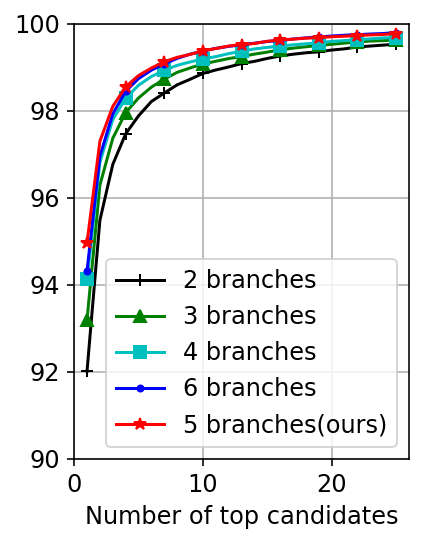}
\end{center}
  \caption{Ablation study on the number of branches in the adaptive receptive field module. As the number of branches increases, average recall goes up to saturation with 5 branches(ours).}
\label{fig:ablation4}
\end{figure}

\begin{table}
\begin{center}
\begin{tabular}{lcc}
\hline
Network & {\em AR@1} & {\em AR@1\%}\\
\hline
1 branch & - & - \\
2 branches & 92.0 & 97.4 \\
3 branches & 93.2 & 97.9 \\
4 branches & 94.1 & 98.3 \\
5 branches (ours) & {\bf 95.0} & {\bf 98.5} \\
6 branches & 94.3 & 98.4 \\
\hline
\end{tabular}
\end{center}
\caption{Ablation study on the number of branches in the adaptive receptive field module. ``-'' means mode collapse. }
\label{tab:ablation5}
\end{table}


We also conduct an experiment on how Average Recall changes with the number of branches in the adaptive receptive field module. As illustrated in Fig.~\ref{fig:ablation4} and Tab.~\ref{tab:ablation5}, as the number of branches increases, the discrimination ability of the descriptors produced by TransLoc3D becomes stronger, and saturation occurs when more than 5 branches are used. The result indicates that the local feature extractor requires aggregating geometric information from a local area larger than a certain threshold to produce sufficiently discriminative descriptors, while oversized receptive fields lead to too complicated geometric information for local descriptors to efficiently represent. More ablation studies can be seen in the appendix.

\subsection{Failure Cases}

\begin{figure}[!t]
\centering
\includegraphics[width=0.8\linewidth]{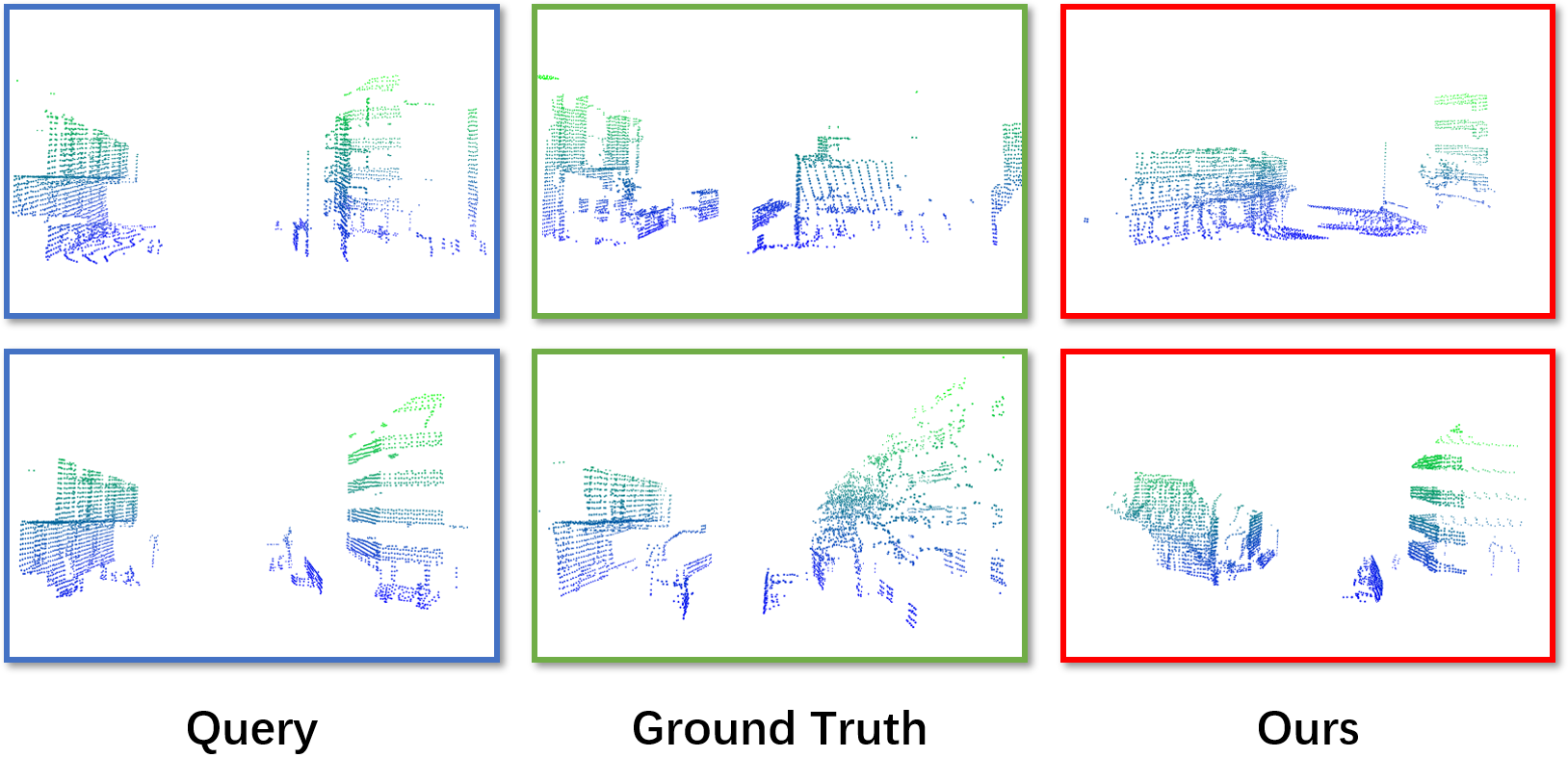}
\caption{Failure cases. We visualize two typical cases in which our proposed model fails.}
\label{fig:failure}
\end{figure}

Although our proposed network has achieved encouraging results on popular datasets, there are still several cases TransLoc3D cannot handle well. Fig.~\ref{fig:failure} illustrates two typical kinds of failure cases. In the first case, geometry of the buildings on different sides of the road changes dramatically, while a similar building (right side of the retrieval result) lies on another location. In the second case, the tree on the street obscures the huge building (right side of the retrieval result) and further changes the distribution of points related to the building, whereas our network does not take occlusion between unmovable objects into consideration.

\section{Conclusion}

In this paper, we put emphasis on three issues in point cloud based place recognition, including moving objects, size difference of the objects and long-range contextual information. We propose TransLoc3D, which combines the advantages of adaptive receptive fields and transformer to tackle these issues. Extensive experiments show that our network achieves state-of-the-art performance on benckmark datasets, and qualitative analysis also demonstrates the effectiveness of our model in complex outdoor scenes. We believe our work can promote further exploitation in visual transformers by utilizing multi-scale geometry information.


\appendix

\section{Network Architecture}

In this section, we provide more details of our method, including sparse voxelization and 3D sparse convolution module.

\subsection{Voxelization}

\begin{figure}[t]
    \centering
    \includegraphics[width=0.5\linewidth]{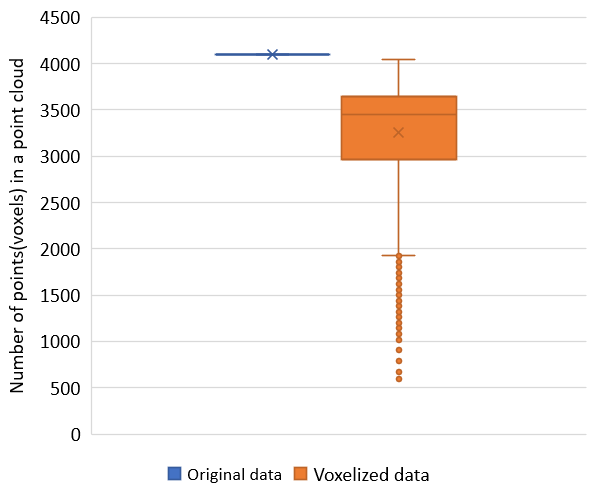}
    \caption{Changes in the number of points during voxelization. About 15\% points on average are removed in the voxelization process, which demonstrates that information loss is quite moderate.}
    \label{fig:number_change}
\end{figure}

\begin{figure}[t]
    \centering
    \includegraphics[width=0.5\linewidth]{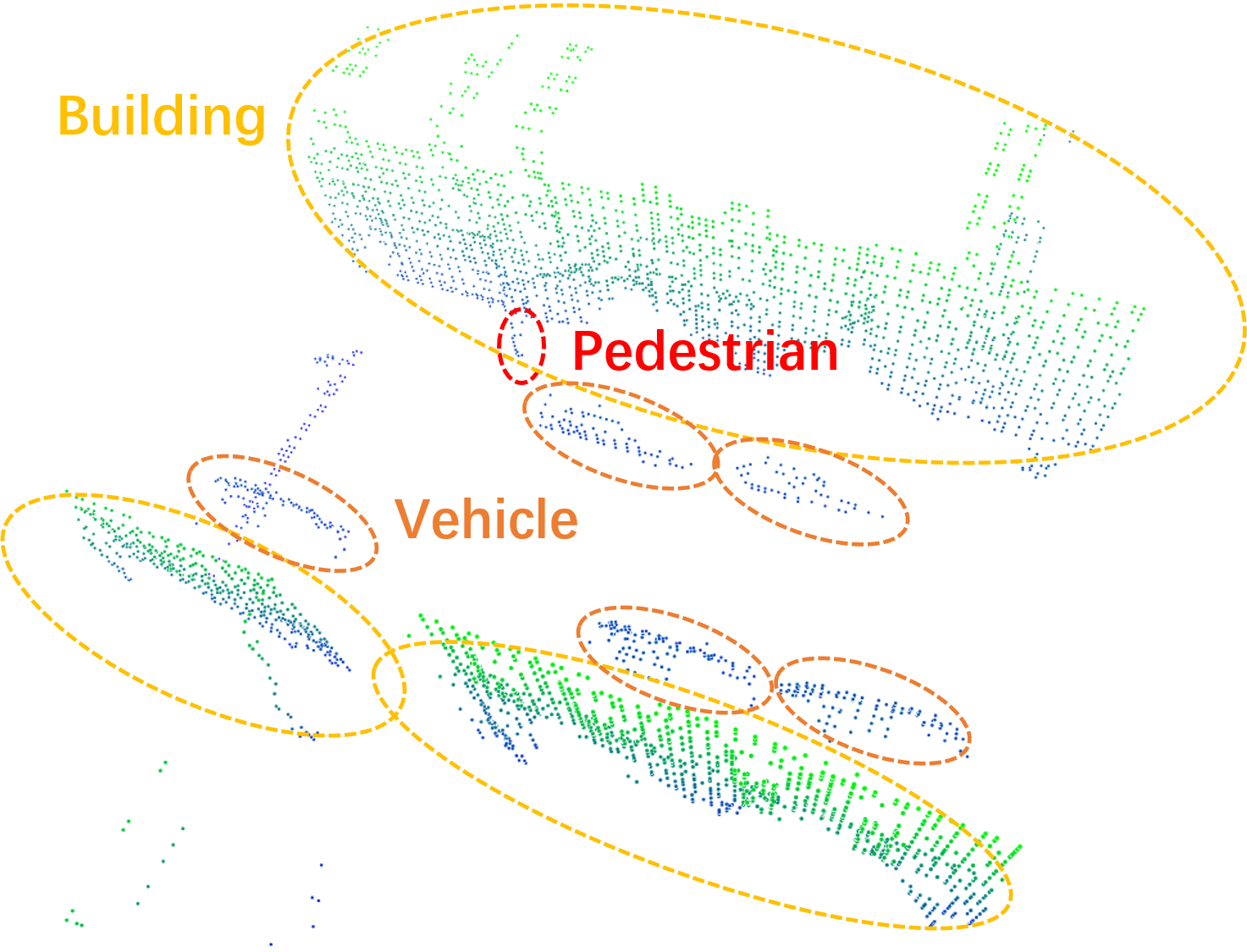}
    \caption{Visualization of voxelized point clouds. 
    }
    \label{fig:visualization}
\end{figure}

To reduce the time and space complexity of obtaining the neighbors of each point, we transform the raw point clouds into sparse voxelization representation before feeding them into our proposed network. However, sparse voxelization inevitably leads to information loss. To quantify information loss introduced in this step, we conduct an experiment on how the number of points changes during voxelization. As illustrated in Fig.~\ref{fig:number_change}, about 600 out of 4096 points are removed on average for being too closed to other points with a quantization size of 0.01. We also visualize the sparse voxel representation in Fig.~\ref{fig:visualization}. As the smallest independent semantic unit, the pedestrian is still roughly recognizable in the complex scene. Thus, we can safely assume that the information loss introduced during voxelization is negligible.

\subsection{3D Sparse Convolution Module}

\begin{table}
\begin{center}
\begin{tabular}{lc}
\hline
Layer & Parameters \\
\hline 
\\[-0.7em]
Conv0 & $\mathbb{C}_{5_k 1_s}^{64}$ \\
\\[-0.7em]
Conv1 & $\mathbb{C}_{2_k 2_s}^{64}$ \\
\\[-0.7em]
\hline
\end{tabular}
\end{center}
\caption{Details of the 3D sparse convolution module.}
\label{tab:conv3d}
\end{table}

We adopt a very shallow 3D CNN consisting of two sparse convolutional layers to aggregate local geometric information and extract point features. The details of the 3D sparse convolution module are shown in Tab.~\ref{tab:conv3d}, where $\mathbb{C}_{a_k b_s}^c$ denotes a convolutional layer with $c$ kernels of shape $a\times a \times a$ and stride $b$. Same as Minkloc3D \cite{komorowski2021minkloc3d}, the first convolution has a $5\times5\times5$ kernel to aggregate information from a large area. In order to extract low-level geometric features, only one $2\times2\times2$ convolution with stride 2 is used in our network, which halves the spatial resolution in each dimension. All convolutions are followed by a batch normalization \cite{ioffe2015batch} layer and a ReLU non-linear activation function.


\section{Experiments}

In this section, we visualize some retrieval results on Oxford RobotCar dataset and provide more ablation studies on hyper-parameter selection. We also conduct a detailed analysis on applying dilated convolutions on sparse feature maps.  

\subsection{Visualization}

\begin{figure}
\begin{center}
   \includegraphics[width=1.0\linewidth]{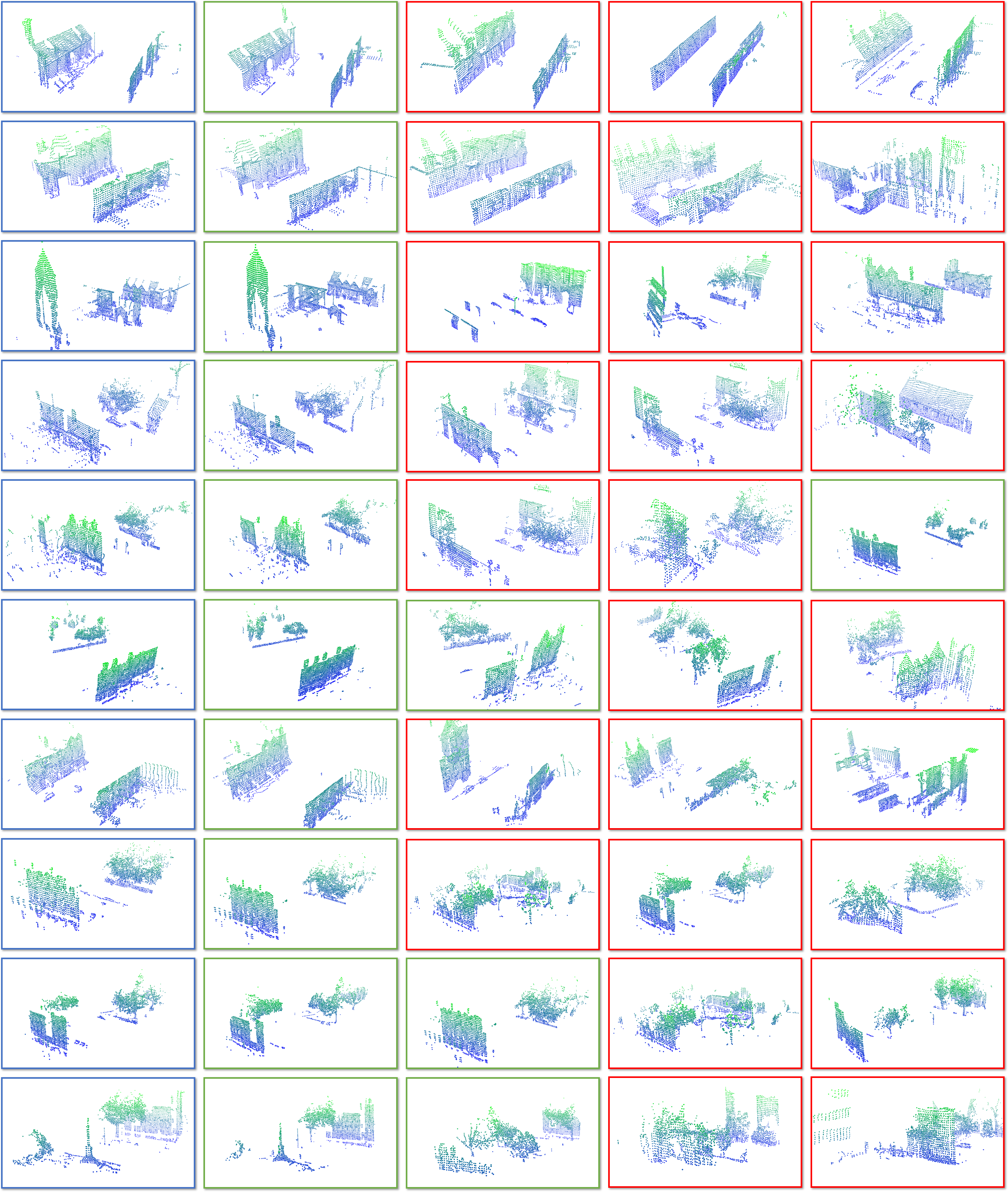}
\end{center}
   \caption{Visualizations of some retrieval results on the Oxford RobotCar dataset. The query point clouds are shown in the leftmost column (blue), and the top 4 nearest neighbors are shown in the other columns in order. The green boxes represent the correct retrieval results, whereas the red boxes indicate the unexpected ones.}
\label{fig:visual2}
\end{figure}

Some retrieval results of our proposed TransLoc3D are shown in Fig.~\ref{fig:visual2}, where the leftmost column (blue) shows the query point clouds and other columns show top 4 nearest neighbors in sequence. The green boxes denote the correct results, while the red ones denote the unexpected retrieval results.

\subsection{Ablation Studies}

\begin{figure}[t]
\begin{center}
   \includegraphics[width=0.5\linewidth]{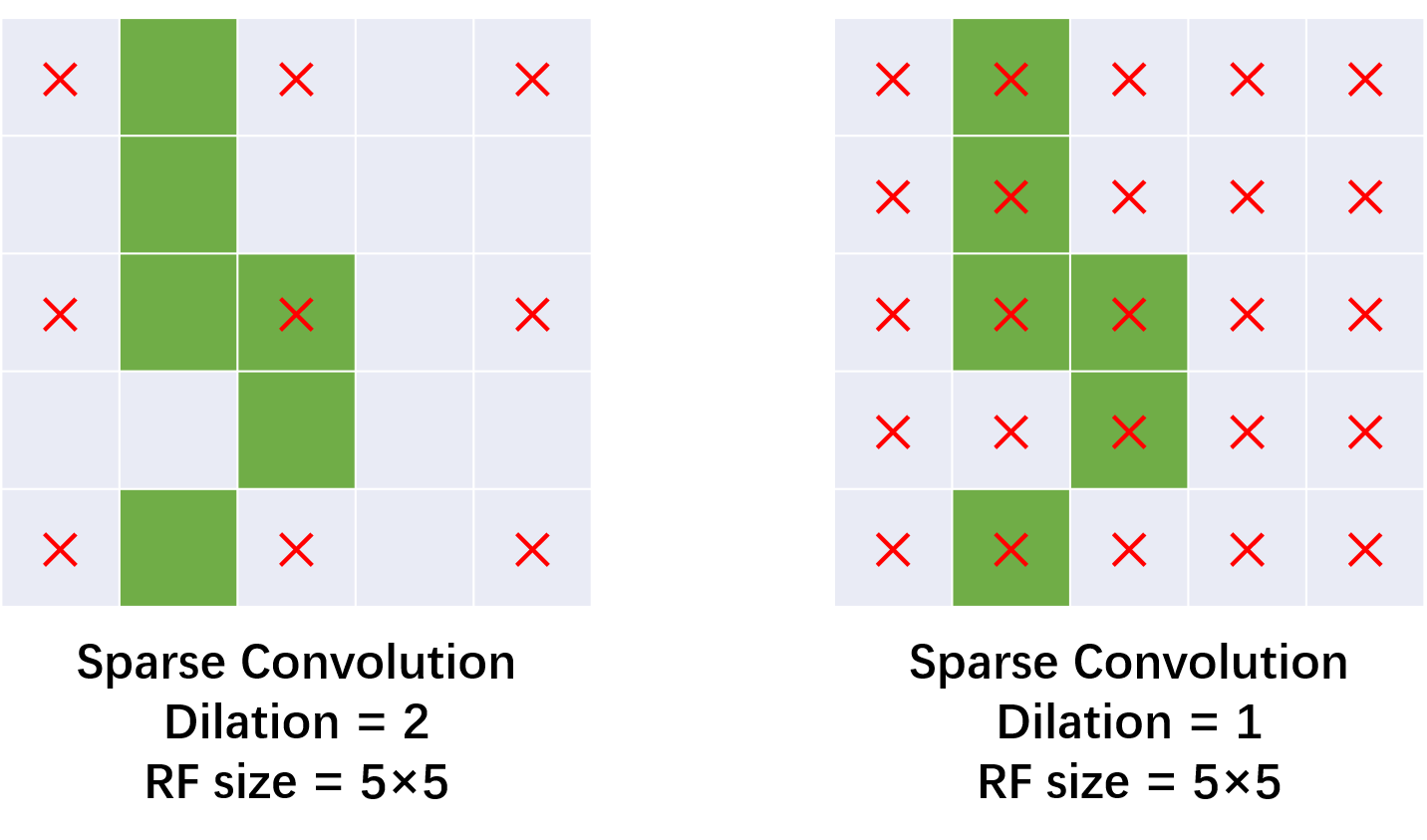}
\end{center}
   \caption{Visualization of dilated convolution and conventional convolution applied on a sparse feature map. Green areas indicate non-zero values in the sparse feature map, while red crosses indicate the receptive field of the central location. Dilated convolution (on the left) cannot aggregate information from the immediate neighbors of the central location.}
\label{fig:dilated_conv}
\end{figure}

\begin{figure}[t]
\begin{center}
   \includegraphics[width=0.3\linewidth]{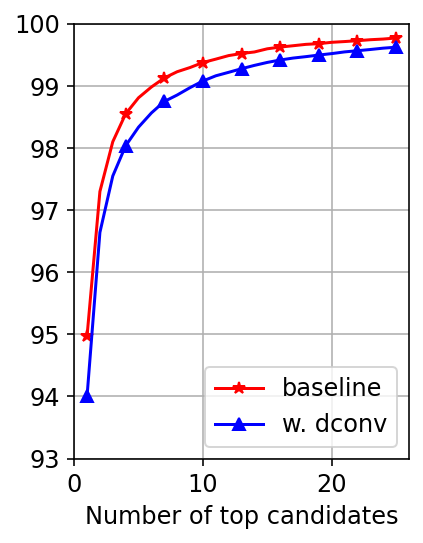}
\end{center}
   \caption{Ablation study on dilated convolution. We find dilated convolution not suitable for sparse feature maps.}
\label{fig:dilated_conv2}
\end{figure}

\begin{figure}[t]
\begin{center}
   \includegraphics[width=0.3\linewidth]{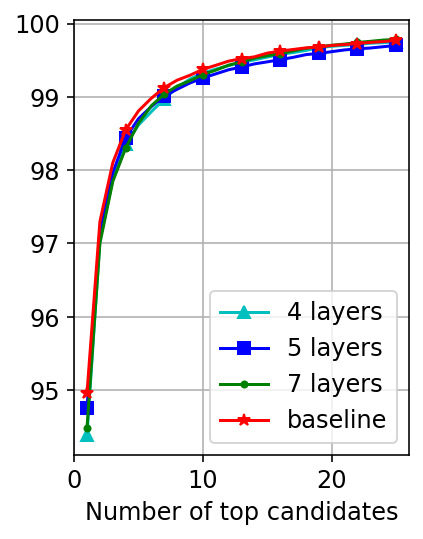}
\end{center}
   \caption{Ablation study on the number of attention layers in the External Transformer. As the number of attention layers increase, average recall goes up to saturation with 6 layers (ours).}
\label{fig:attn_layers}
\end{figure}

Different from SK-Net~\cite{li2019selective} which adopts dilated convolutions for further improving efficiency, we build our adaptive receptive field module with conventional convolutions. We argue that dilated convolution is not suitable for sparse feature maps and provide qualitative and quantitative analysis to support this claim. 
Notably, the positions of nonzero features in sparse feature maps do not change when adopting sparse convolution, and most non-zero features extracted from point cloud data are distributed along the outer surface of objects. We observe that dilated convolution applied on the sparse feature map may aggregate much less local information than conventional convolution. Without loss of generality, we visualize the feature map of a 2D point cloud as illustrated in Fig.~\ref{fig:dilated_conv}, where nonzero values (green) are evenly scattered over the left surface or right surface of the 2D object. The number of non-zero features fed to dilated convolutional layer is lower than the theoretical maximum, and as a result the dilated convolutions cannot aggregate local information effectively. Fig.~\ref{fig:dilated_conv2} shows the quantitative comparison between adaptive receptive field module built on conventional convolutions and dilated convolutions. It can be seen that the dilated convolution-based network has obvious degradation on average recall@N, which demonstrates that dilated convolution applied on sparse feature maps has negative impact on performance.

We also provide an analysis on the number of attention layers in the External Transformer. As illustrated in Fig.~\ref{fig:attn_layers}, TransLoc3D with 6 layers achieves the best results. However, it only has a minor improvement on average recall compared with other settings, which demonstrates the robustness of TransLoc3D to hyper-parameter selection.

\section*{Acknowledgements}

The authors thank Meng-Hao Guo and Yu He for their continued support and fruitful discussions. This work was supported by the National Key Technology R\&D Program (Project Number 2017YFB1002604), the National Natural Science Foundation of China (Project Numbers 61772298, 61832016), Key Research Projects of the Foundation Strengthening Program under Grant No. 2020JCJQZD01412, Research Grant of Beijing Higher Institution Engineering Research Center, and Tsinghua–Tencent Joint Laboratory for Internet Innovation Technology.

\bibliographystyle{plain}
\bibliography{main}

\address{Tsinghua University\\
China\\
\email{xutx21@mails.tsinghua.edu.cn}}

\address{Tsinghua University\\
China\\
\email{guoyc19@mails.tsinghua.edu.cn}}

\address{National Defence University of PLA\\
China\\
\email{2683014096@qq.com}}

\address{Chinese Academy of Sciences\\
China\\
\email{yuge@csu.ac.cn}}

\address{Cardiff University\\
             United Kingdom\\
\email{LaiY4@cardiff.ac.uk}}

\address{Tsinghua University\\
             China\\
\email{shz@tsinghua.edu.cn}\\
\received{February 26, 2022}\\
\accepted{August 9, 2022}}

\end{document}